\documentclass{article}

\PassOptionsToPackage{numbers,compress}{natbib}
\usepackage[preprint]{neurips_2026}

\usepackage{algorithm}
\usepackage{algorithmicx}
\usepackage{algpseudocode}
\usepackage{enumitem}
\usepackage[utf8]{inputenc}
\usepackage[T1]{fontenc}
\usepackage{hyperref}
\usepackage{url}
\usepackage{booktabs}
\usepackage{amsfonts}
\usepackage{nicefrac}
\usepackage{microtype}
\usepackage{xcolor}
\usepackage{graphicx}
\usepackage{amsmath}
\usepackage{multirow}

\title{Rein3D: Reinforced 3D Indoor Scene Generation with Panoramic Video Diffusion Models}

\author{%
\vspace{-0.8cm}
\\
\textbf{Dehui Wang}$^{1}$ \quad
\textbf{Rong Wei}$^{2}$ \quad
\textbf{Yue Shi}$^{1}$ \quad
\textbf{Congsheng Xu}$^{1}$ \quad
\textbf{Shoufa Chen}$^{4}$ \\
\textbf{Dingxiang Luo}$^{1}$ \quad
\textbf{Tianshuo Yang}$^{4}$ \quad
\textbf{Xiaokang Yang}$^{1}$ \quad
\textbf{Wei Sui}$^{3}$ \quad
\textbf{Yusen Qin}$^{3}$ \\
\textbf{Rui Tang}$^{2}$ \quad
\textbf{Yao Mu}$^{1}$\thanks{Corresponding Author. Email: \texttt{muyao@sjtu.edu.cn}} \\
\\
$^1$Shanghai Jiao Tong University \quad
$^2$Manycore Tech Inc. \quad
$^3$D-Robotics \\
$^4$The University of Hong Kong \\
}

\begin{document}

\maketitle

\begin{abstract}
Reconstructing complete and explorable 3D indoor scenes from a single panorama is intrinsically ill-posed, because camera translation reveals large disoccluded regions that are unobserved from the panorama center and difficult to recover with globally consistent geometry. We present Rein3D, a restore-and-refine framework that formulates single-panorama 3D scene completion as 3D-prior-guided panoramic video restoration followed by global 3D Gaussian Splatting refinement. Given a panoramic image and its depth map, either provided or estimated, Rein3D first lifts the RGB-D panorama into a coarse 3DGS representation, and then performs radial camera exploration to render RGB-Alpha panoramic videos that expose holes, disocclusions, incomplete geometry, and opacity artifacts caused by single-view initialization. Instead of hallucinating independent novel views, Rein3D restores these degraded panoramic sequences with a panoramic Video-to-Video diffusion model, where the Context Anchor and spherical adaptation help preserve scene consistency under large camera motion. The restored videos are then used as pseudo observations to refine a single global 3DGS, yielding a more complete and view-consistent indoor scene. To support this formulation, we construct PanoV2V-15K, a large-scale paired clean--degraded 360-degree video dataset with 15,050 indoor scenes for panoramic video restoration. Extensive evaluations on PanoV2V-15K, out-of-distribution 3D-FRONT, and Structured3D show that Rein3D improves both panoramic restoration and final 3D scene reconstruction, reducing FVD from 14.13 to 9.82 on OOD restoration and improving ground-truth novel-view PSNR from 15.05 to 17.64 under single-panorama 3D reconstruction.
\end{abstract}

    \begin{figure}[t]
\vspace{-5pt}
        \centering
\includegraphics[width=0.85\linewidth]{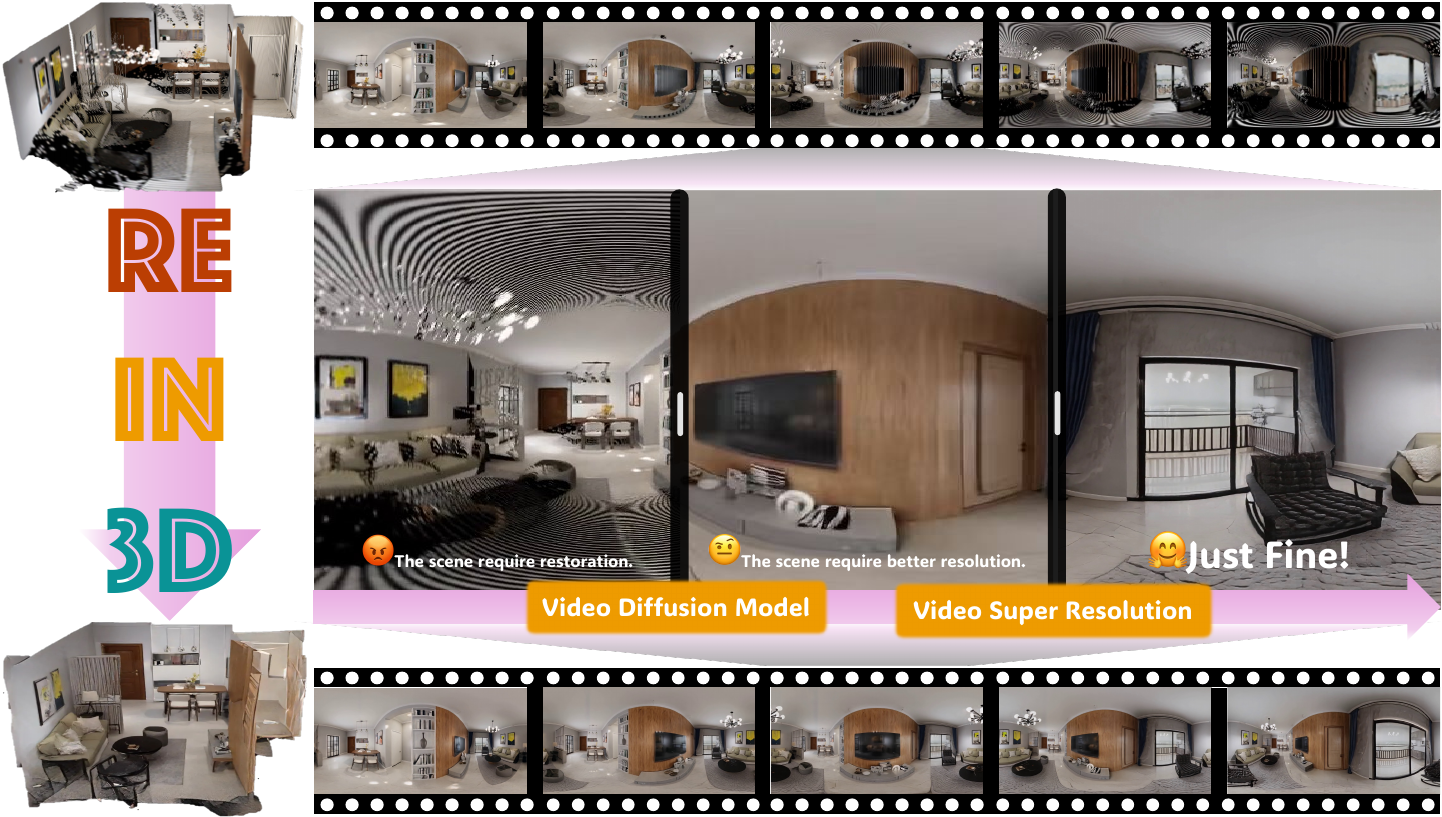}
        \caption{Overview of Rein3D. Starting from a single RGB-D panorama, Rein3D initializes a coarse 3DGS scene and renders degraded panoramic videos along radial trajectories to expose unseen regions.
A panoramic video diffusion model restores these videos with temporally consistent priors, and the restored views are fused to refine the global 3DGS.
This restore-and-refine paradigm enables photorealistic and globally consistent indoor scene reconstruction under large viewpoint changes.
\vspace{-5pt}
}
        
        \label{fig:teaser}
    \end{figure}

\section{Introduction}
\label{sec:intro}
Synthesizing high-quality 3D indoor scenes from sparse inputs presents promising application prospects, particularly in advancing VR/AR and simulation platforms for Embodied AI~\cite{yang2024physcene,miao2025towards}.
In this work, we focus on a practical and informative sparse input: a single panoramic image, optionally paired with a given or estimated depth map.
Compared with perspective images, a panorama captures the full 360$^\circ$ appearance around the camera center and provides rich global context for indoor scene modeling.
However, reconstructing a complete and explorable 3D scene from a single panorama remains fundamentally ill-posed:
the input only observes the scene from one center location and cannot directly reveal disoccluded regions that appear when the camera moves away.
Consequently, inferring massive amounts of missing information for unseen areas while maintaining geometric consistency remains a significant challenge.

Deep generative models, particularly diffusion models~\cite{ho2020denoising,song2020score,ramesh2022hierarchical,rombach2022high,huang2025omnix}, address this by leveraging strong 2D visual priors.
However, standard image-based methods~\cite{yu2024wonderjourney,yu2024viewcrafter,yu2025wonderworld,yang2024scenecraft,ling2025scenethesis} often suffer from accumulated geometric errors when their generated views are fused into a global 3D representation.
While strategies like explicit constraints or multi-view synthesis~\cite{tang2024diffuscene,zhou2024dreamscene360,fang2025spatialgen} alleviate this issue, they remain computationally intensive and operationally cumbersome.
In contrast, video diffusion methods~\cite{he2024cameractrl,chen2025flexworld,wu2025video,chen2024gentron}, especially Video-to-Video (V2V) approaches guided by 3D priors, have demonstrated impressive results in producing temporally coherent novel observations.
Nevertheless, these methods typically operate with a limited Field-of-View (FoV).
Covering an entire indoor scene requires stitching many views along carefully designed camera trajectories, which is computationally expensive and often yields weaker global consistency than panoramic representations.

Panoramic video diffusion~\cite{wang2024360dvd,fang2025panoramic,xia2025panowan} offers a natural solution by modeling the full 360$^\circ$ view at each camera location, but it faces a critical data bottleneck for 3D scene restoration.
To bridge this gap, we construct a large-scale dataset \textbf{PanoV2V-15K} for panoramic video restoration.
This dataset comprises over 15,000 distinct indoor scenes, consisting of paired videos: low-quality panoramic renderings with artifacts and their corresponding clean ground truths.
These degraded videos are generated from coarse 3D representations initialized from single panoramic RGB-D observations, and therefore reflect the typical holes, distortions, and missing regions encountered during single-pano 3D reconstruction.
Leveraging the 360-degree field of view, we simplify the data generation by employing pure linear trajectories.
Since the panorama captures everything around the camera, this simple motion is sufficient to expose large disoccluded regions, avoiding the complex path planning required by limited-FoV approaches~\cite{he2024cameractrl,wang2024motionctrl,huang2025voyager}.

As shown in Fig.~\ref{fig:teaser}, building on our newly constructed dataset, we propose \textbf{Rein3D}, a novel framework for high-fidelity 3D indoor scene reconstruction from a single panorama.
Specifically, our pipeline consists of three key steps.
First, given a panoramic image and its depth map, either provided or estimated, we lift the RGB-D panorama into an initial 3D Gaussian Splatting (3DGS)~\cite{kerbl20233d} representation.
Second, to efficiently reveal and restore unseen regions, we design a robust radial exploration strategy.
Starting from the panorama center, the camera moves outwards along uniformly distributed trajectories to capture imperfect panoramic renderings from the coarse 3DGS.
These degraded panoramic sequences are then repaired by a Video-to-Video diffusion model and enhanced via Video Super-Resolution.
Finally, the restored high-fidelity videos are projected back as pseudo observations to optimize the global 3DGS, resulting in a complete and high-quality 3D indoor scene.

We highlight the key contributions of our work as follows:
\vspace{-0.6em}
\begin{itemize}[leftmargin=*,itemsep=2pt, topsep=2pt, parsep=0pt]
    \item We introduce PanoV2V-15K, a large-scale dataset of paired clean--degraded 360$^\circ$ videos, enabling diffusion-based panoramic restoration for downstream single-pano 3D scene reconstruction.
    \item We propose Rein3D, a novel framework that formulates single-panorama 3D scene completion as 3D-prior-guided panoramic video restoration followed by global 3DGS refinement.
    \item Extensive experiments demonstrate our superior performance in both panoramic video restoration and 3D indoor scene reconstruction, particularly for large-scale environments and long-range exploration.
\end{itemize}

\section{Related Work}

\subsection{3D Indoor Scene Generation}
Powered by advances in differentiable rendering
\cite{mildenhall2021nerf,kerbl20233d} , 3D generative models have expanded from isolated objects \cite{wang2023prolificdreamer,xiang2025structured} to complex scene synthesis\cite{bahmani2025lyra,li2025flashworld}.
Early iterative methods employ a ``warp-and-refine'' paradigm to extend the field of view \cite{wiles2020synsin,hollein2023text2room,chung2023luciddreamer,meng2025scenegen} but often suffer from geometric drift over long trajectories. 
To ensure global consistency, subsequent works utilize holistic 360-degree panoramas \cite{pu2024pano2room} or layouts
\cite{fang2025spatialgen} as unified priors, lifting them into 3D Gaussian Splatting for real-time rendering \cite{hu2024scenecraft,wang2024perf,zhou2024dreamscene360}. 
However, these panorama-centric approaches struggle with significant camera translations due to occlusion. 
Consequently, recent methods \cite{yu2024viewcrafter,sun2024dimensionx,huang2025voyager} use video diffusion models to hallucinate consistent details during movement. 
In this work, we propose a novel framework that initializes a globally consistent scene from a panorama and leverages video diffusion models to coherently restore and refine scene details during zoom-in exploration.

\subsection{Panoramic Video Diffusion Models}
The evolution of diffusion models \cite{ho2020denoising,song2020denoising} revolutionized video synthesis, transitioning from UNet-based architectures with temporal layers \cite{he2022latent,blattmann2023stable} to scalable Diffusion Transformers \cite{peebles2023scalable} that master complex spatio-temporal dynamics \cite{hacohen2024ltx,kong2024hunyuanvideo,yang2024cogvideox,chen2025goku,wan2025wan}. 
Although these foundation models excel at perspective content, applying them to 360-degree videos remains challenging due to spherical distortion and boundary discontinuity.
Pioneering efforts like 360DVD \cite{wang2024360dvd} and VideoPanda \cite{xie2025videopanda} employ adapters or multi-view attention to handle spherical geometry.
Meanwhile, specific applications focus on animating static panoramas \cite{li20244k4dgen} or expanding narrow field-of-view videos to full immersive views \cite{tan2024imagine360,ma2024vidpanos}.
Rencently, ViewPoint \cite{fang2025panoramic} introduces a novel map representation combined with pano-perspective attention to ensure spatial continuity, whereas PanoWan \cite{xia2025panowan} employs latitude-aware sampling and rotated denoising to lift powerful DiT backbones to the panoramic domain.
In this work, we propose a video-to-video framework that utilizes these advanced DiT priors for panoramic scene restoration.
By conditioning on preliminary renderings, our method effectively recovers high-fidelity details in occluded regions while ensuring temporal consistency.

\section{PanoV2V-15K Dataset Construction}
\label{sec:dataset_construction}

To train and evaluate our panoramic video restoration model, we construct \textbf{PanoV2V-15K}, a large-scale paired clean--degraded panoramic video dataset following the data construction protocol of prior work~\cite{fang2025spatialgen}. 
The dataset contains 15,050 indoor scenes with diverse room types, layouts, and visual appearances, synthesized from a curated collection of professionally designed indoor environments.

\begin{figure}[t]
    \centering
    \includegraphics[width=0.8\linewidth]{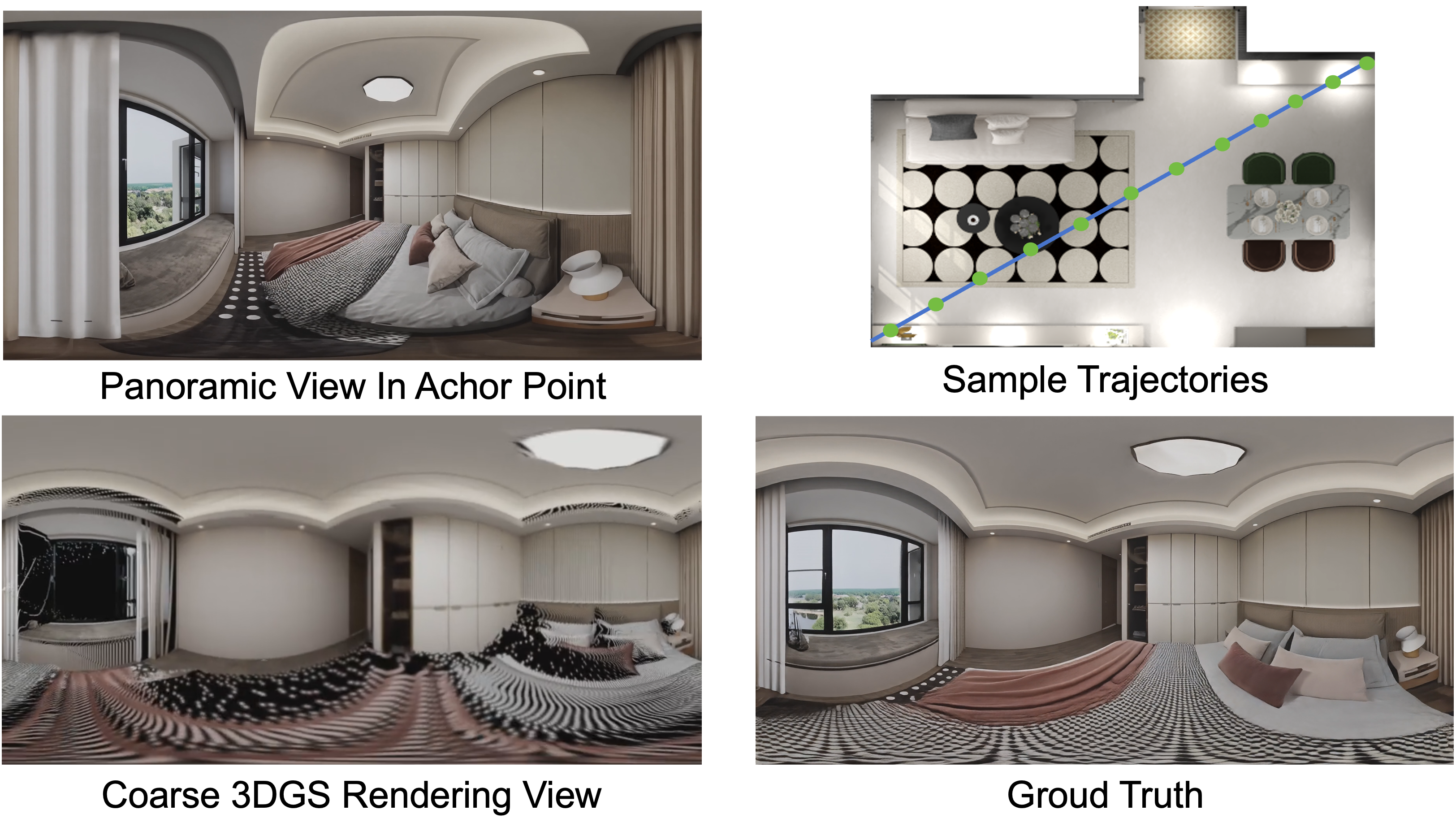}
    \caption{
    Illustration of PanoV2V-15K construction.
    For each scene, we render a clean panoramic video along the longest traversable trajectory and generate its paired degraded RGB-alpha video by rendering a coarse 3DGS initialized from the first panoramic RGB-D observation.
    }
    \label{fig:dataset_construction}
\end{figure}

For each room, we select the longest traversable diagonal trajectory on the $XY$ plane, which provides large camera displacement and exposes disoccluded regions during motion. 
A virtual omnidirectional camera is placed at a fixed height of 1.6\,m and moves along this trajectory at 1\,m/s, recording a clean $360^\circ$ panoramic video at 10 FPS. 
This yields temporally coherent ground-truth videos with smooth linear camera motion, as illustrated in Fig.~\ref{fig:dataset_construction}.

To construct the paired degraded video, we use only the first panoramic RGB-D observation of each trajectory. 
The first clean panorama and its ground-truth depth map are lifted into a coarse 3DGS representation by unprojecting each equirectangular pixel into 3D and initializing a Gaussian primitive with its corresponding position and color. 
The remaining Gaussian attributes are initialized deterministically with isotropic scales, identity rotations, and high opacity, without iterative optimization.

We then render this coarse 3DGS along the same trajectory as the clean video. 
Since the coarse 3DGS is initialized from only a single panoramic RGB-D observation, its novel-view renderings naturally contain holes, disocclusions, incomplete geometry, and opacity artifacts. 
These RGB-alpha panoramic renderings form the degraded conditioning video, while the original clean panoramic video serves as the ground truth. 
Thus, each sample in PanoV2V-15K consists of a paired clean--degraded panoramic video, together with the first panorama as the context anchor.

\section{Method}
\label{sec:method}
\begin{figure*}[htbp] 
    \centering
    \includegraphics[width=1.0\linewidth]{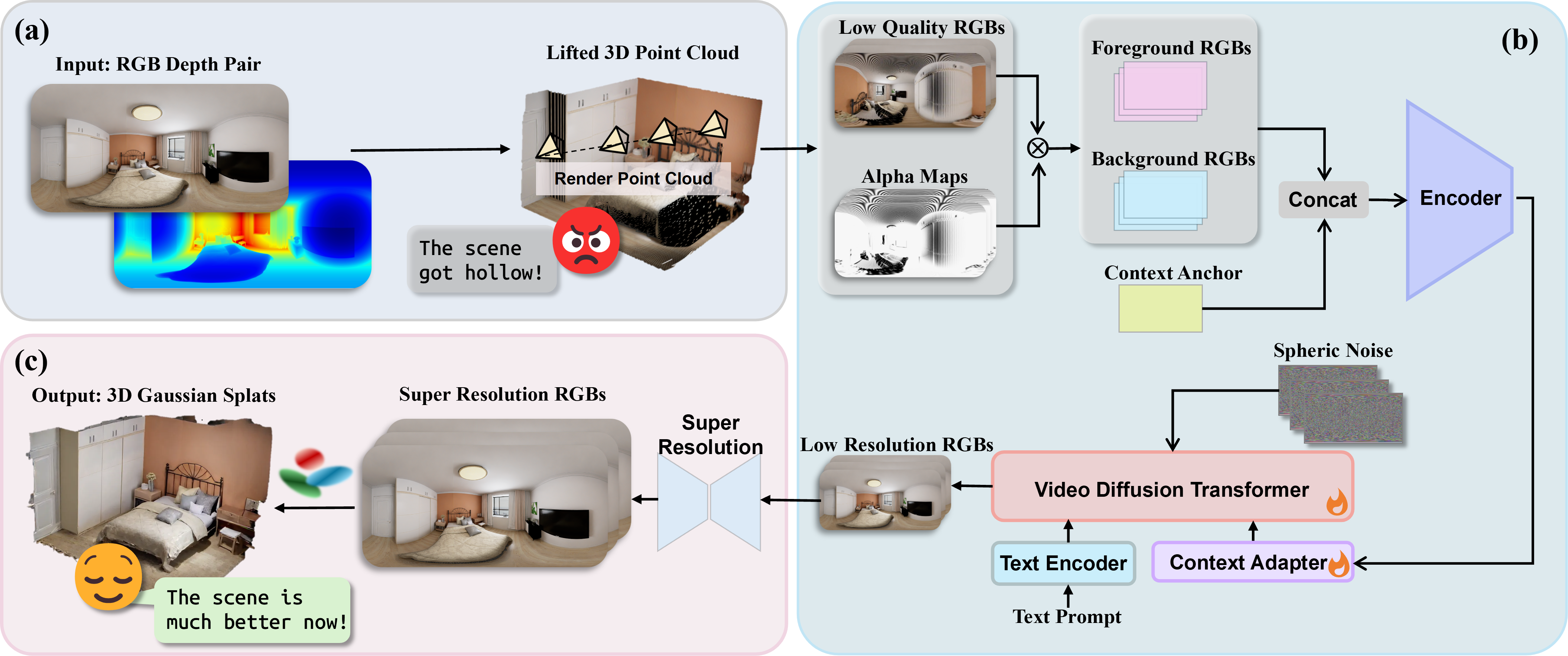}
    \caption{
    Overview of the Rein3D pipeline.
    \textbf{(a)} Given a single panoramic RGB-D observation, we lift it into an initial 3D point cloud / coarse 3DGS representation. Rendering this coarse representation from novel viewpoints exposes holes, missing geometry, and opacity artifacts caused by single-view initialization.
    \textbf{(b)} The resulting low-quality RGB-alpha panoramic videos are restored by our panoramic video diffusion model. The RGB-alpha inputs are decomposed into observed and missing-region streams, combined with a context anchor, and processed with latitude-aware noise sampling to produce temporally consistent restored panoramas.
    \textbf{(c)} The restored videos are further enhanced by super-resolution and fused back as pseudo observations to refine the global 3DGS, yielding a more complete and view-consistent indoor scene.
    }
    \label{fig:pipeline}
\end{figure*}

As illustrated in Fig.~\ref{fig:pipeline}, Rein3D reconstructs an explorable indoor 3D scene from a single panoramic RGB-D observation via a restore-and-refine paradigm.
We first lift the input panorama and its provided or estimated depth into a coarse 3DGS representation, then perform radial exploration to render degraded RGB-alpha panoramic videos that expose disocclusions and rendering artifacts.
A panoramic video-to-video diffusion model restores these videos with temporally consistent priors, and the restored frames are further enhanced by video super-resolution.
Finally, we project the high-fidelity panoramic frames into perspective views and use them as pseudo observations to refine the global 3DGS.
We describe the initialization and exploration in Sec.~\ref{sec:init}, the restoration network in Sec.~\ref{sec:network}, and the final 3D refinement in Sec.~\ref{sec:refine}.

\subsection{Initialization and Camera Exploration}
\label{sec:init}

\noindent \textbf{Coarse 3DGS Initialization.}
Given an input panorama $\mathbf{I}_0$ and its associated depth map $\mathbf{D}_0$, we first lift the panoramic RGB-D observation into a coarse 3DGS representation.
For each equirectangular pixel, we compute its spherical viewing direction and back-project it using the corresponding depth value.
The resulting 3D point is used as the center of a Gaussian primitive, with its color initialized from the associated RGB pixel.
We initialize the remaining Gaussian attributes with isotropic scales, identity rotations, and high opacity, producing a coarse but renderable 3D prior without iterative optimization.
Although this representation only contains surfaces visible from the panorama center, it provides RGB-alpha renderings under novel camera poses.

\noindent \textbf{Radial Trajectory Generation.}
To expose regions that are missing or under-constrained in the coarse 3DGS, we employ a simple radial trajectory strategy.
Using the panoramic depth, we estimate a safe radial depth profile for camera motion.
Instead of relying on small local perturbations around the input viewpoint, we sample $\tau$ uniformly distributed trajectories radiating outward from the panorama center and search for a global orientation that maximizes the cumulative safe travel distance.
We then render degraded RGB-alpha panoramic videos from the coarse 3DGS along these trajectories.
These videos naturally contain disocclusion holes, incomplete geometry, and opacity artifacts, and serve as the degraded inputs for our restoration network.
Additional details of the trajectory generation procedure are provided in Appendix~\ref{app:camera_exploration}.

\subsection{Panoramic Video Restoration Network}
\label{sec:network}

To transform imperfect RGB-Alpha renderings into high-fidelity videos, we design a specialized restoration network based on the Wan2.1-1.3B \cite{wan2025wan} DiT backbone.
We adapt this foundation model to the panoramic domain through a conditional injection mechanism and specific spherical optimizations.

\noindent \textbf{Architecture and Conditioning.}
Inspired by VACE \cite{vace}, we utilize a context adapter architecture to effectively handle multi-modal inputs.
To ensure global consistency during camera exploration, we explicitly incorporate a Context Anchor mechanism into the conditioning signal.
Specifically, the input RGB-Alpha video is processed to distinguish between observed content and missing regions. 
Let $\mathbf{V} \in \mathbb{R}^{T \times 3 \times H \times W}$ denote the RGB video and $\mathbf{M} \in [0, 1]^{T \times 1 \times H \times W}$ denote the alpha channel. 
We decompose the input into two complementary streams:
\begin{equation}
    \mathbf{V}_{\text{background}} = \mathbf{V} \odot (\mathbf{1} - \mathbf{M}), \quad \mathbf{V}_{\text{foreground}} = \mathbf{V} \odot \mathbf{M}
\end{equation}
These streams are independently encoded by the pre-trained VAE encoder $\mathcal{E}$ and concatenated along the channel dimension to form the video latent features $\mathbf{Z}_{\text{video}}$:
\begin{equation}
    \mathbf{Z}_{\text{video}} = \text{Concat}(\mathcal{E}(\mathbf{V}_{\text{foreground}}), \mathcal{E}(\mathbf{V}_{\text{background}}))
\end{equation}
Crucially, since our trajectory begins from the scene origin, the starting panoramic view $\mathbf{I}_0$ is fully observable (i.e., $\mathbf{M}_0 = \mathbf{1}$). 
We treat $\mathbf{I}_0$ as the \textbf{Context Anchor}, encoding it to obtain $\mathbf{Z}_{anchor} = \mathcal{E}(\mathbf{I}_0)$. 
Finally, we construct the input condition $\mathbf{Z}_{condition}$ by prepending the anchor to the video features along the temporal dimension:
\begin{equation}
    \mathbf{Z}_{\text{condition}} = [\mathbf{Z}_{\text{anchor}}, \mathbf{Z}_{\text{video}}]
\end{equation}
This design ensures that the generation process is explicitly grounded in the initial scene geometry, preventing temporal drift and color inconsistency.

\noindent \textbf{Spherical Adaptation.}
Standard diffusion models treat frames as flat images, while our inputs are equirectangular panoramas with latitude-dependent distortion.
Pixels near the poles are horizontally stretched and over-represented relative to their physical area on the sphere.
We therefore introduce two spherical adaptations for panoramic video restoration.

\noindent (1) \textbf{Latitude-aware Noise Sampling.}
Following PanoWan~\cite{xia2025panowan}, we adapt the initial noise sampling to the spherical distortion of ERP images.
Given an input resolution of $H \times W$, we define the latitude $\phi$ for a pixel at vertical coordinate $y$ as $\phi = (0.5 - y/H)\pi$.
We warp the horizontal noise coordinate as
\begin{equation}
    x' = x_c + (x - x_c) \cdot \cos(\phi),
\end{equation}
where $x_c$ is the horizontal image center.
This latitude-dependent coordinate warp adjusts the horizontal noise distribution to better match the non-uniform sampling density of ERP images.

\noindent (2) \textbf{Latitude-decay Loss.}
Complementary to noise sampling, we reweight the training objective according to the physical area represented by each ERP pixel.
Since polar pixels correspond to smaller surface areas on the sphere, we reduce their contribution using a latitude-dependent weight:
\begin{equation}
    W(\phi) = \lambda + (1 - \lambda) \cos(\phi),
\end{equation}
where $\lambda$ prevents polar regions from being completely ignored.
This loss reweighting reduces the dominance of over-sampled polar pixels while preserving valid learning signals across the entire panorama.

\subsection{3D Scene Refinement}
\label{sec:refine}
To enhance visual quality of videos, we first employ FlashVSR~\cite{zhuang2025flashvsr}
to upsample the video sequences, effectively recovering high-frequency details.
These high-resolution panoramic frames are then projected into perspective views to serve as pseudo-ground truths.
Finally, we fine-tune the 3D Gaussian Splatting against these views. 
During this optimization, we incorporate robust densification strategies and anti-aliasing techniques to effectively suppress geometric artifacts and further refine the scene texture.

\section{Experiments}
\subsection{Implementation Details}
We implement our Rein3D framework using PyTorch.
The video restoration network is initialized from the pre-trained VACE-1.3B~\cite{vace} checkpoint.
We fine-tune the model on our proposed dataset using the AdamW~\cite{loshchilov2017decoupled} optimizer with a constant learning rate of $5 \times 10^{-5}$.
During training, the input panoramic video sequences are fixed to a length of 41 frames, with a spatial resolution of $448 \times 896$.
The total batch size is set to 32, and the training proceeds for 5,000 steps on 4 NVIDIA H200 (141GB) GPUs.

For the final 3D scene refinement stage, we utilize the efficient gsplat library~\cite{ye2025gsplat}. We optimize the 3D Gaussian Splatting for 15,000 steps against the upsampled pseudo-ground truths. To ensure high rendering quality, we enable anti-aliasing \cite{yu2024mip} during rasterization to effectively suppress geometric artifacts.

\subsection{Comparison on Video Restoration}

We compare our method with ProPainter~\cite{zhou2023propainter}, a transformer-based video inpainting model, and VACE~\cite{vace}, a state-of-the-art foundation model for video generation and editing.
We also fine-tune VACE on the PanoV2V-15K training set, denoted as VACE-FT, while ProPainter uses its official pre-trained weights.

\textbf{Datasets and metrics.}
Since no existing benchmark directly targets RGB-alpha conditioned panoramic video restoration, we evaluate on two test sets.
The first contains 50 clips randomly selected from PanoV2V-15K for in-domain evaluation.
To further test out-of-distribution generalization, we construct another 50 paired clean--degraded panoramic videos from 3D-FRONT~\cite{fu20213d} using the same data construction protocol.
Both our model and VACE-FT are trained only on PanoV2V-15K and are directly evaluated on 3D-FRONT without further adaptation.

We report PSNR and SSIM~\cite{wang2004image} for reconstruction quality, FVD~\cite{unterthiner2019fvd} for temporal consistency, and WS-PSNR / WS-SSIM for spherical reconstruction fidelity under equirectangular projection.

\begin{table*}[htbp]
\centering
\footnotesize
\caption{
Quantitative comparison on panoramic video restoration.
VACE-FT denotes VACE fine-tuned on PanoV2V-15K.
}
\label{tab:restoration_comparison}
\resizebox{0.8\textwidth}{!}{
\begin{tabular}{c l c c c c c}
\toprule
Dataset & Method & PSNR $\uparrow$ & SSIM $\uparrow$ & WS-PSNR $\uparrow$ & WS-SSIM $\uparrow$ & FVD $\downarrow$ \\
\midrule
\multirow{4}{*}{PanoV2V-15K}
& ProPainter & 18.46 & 0.760 & 17.79 & 0.740 & 14.56 \\
& VACE & 21.17 & 0.787 & 20.80 & 0.771 & 11.67 \\
& VACE-FT & 22.18 & 0.798 & 21.89 & 0.790 & 9.27 \\
& \textbf{Ours} & \textbf{23.77} & \textbf{0.823} & \textbf{23.45} & \textbf{0.820} & \textbf{5.26} \\
\midrule
\multirow{4}{*}{3D-FRONT}
& ProPainter & 19.12 & 0.674 & 18.68 & 0.656 & 29.27 \\
& VACE & 21.91 & 0.701 & 21.00 & 0.702 & 25.50 \\
& VACE-FT & 22.55 & 0.775 & 21.50 & 0.756 & 14.13 \\
& \textbf{Ours} & \textbf{24.70} & \textbf{0.808} & \textbf{24.35} & \textbf{0.804} & \textbf{9.82} \\
\bottomrule
\end{tabular}
}
\end{table*}

\textbf{Results.}
As shown in Tab.~\ref{tab:restoration_comparison}, our method consistently outperforms ProPainter, VACE, and VACE-FT across all metrics on both test sets.
The improvements on PanoV2V-15K demonstrate stronger in-domain restoration quality, while the gains on 3D-FRONT show that the learned panoramic restoration model generalizes well to unseen scene distributions.
The consistent advantages in WS-PSNR and WS-SSIM further confirm that our model better handles the spherical geometry of equirectangular videos, and the lower FVD indicates improved temporal consistency.

\subsection{Comparison on 3D Scene Reconstruction}
\label{sec:scene_comparison}

We evaluate Rein3D for single-panorama 3D indoor scene reconstruction.
Given one panoramic RGB-D observation, each method reconstructs an explorable 3D scene, from which we render novel views for evaluation.
We conduct two complementary evaluations.
First, we evaluate no-reference visual quality on Structured3D~\cite{zheng2020structured3d}, where the goal is to assess the perceptual quality and global consistency of rendered views under large viewpoint changes.
Second, we evaluate novel view synthesis accuracy on 3D-FRONT~\cite{fu20213d}, where ground-truth target views are available for direct comparison.

\subsubsection{No-reference Evaluation on Structured3D.}

\begin{figure*}[htbp]
    \centering
    \includegraphics[width=0.85\linewidth]{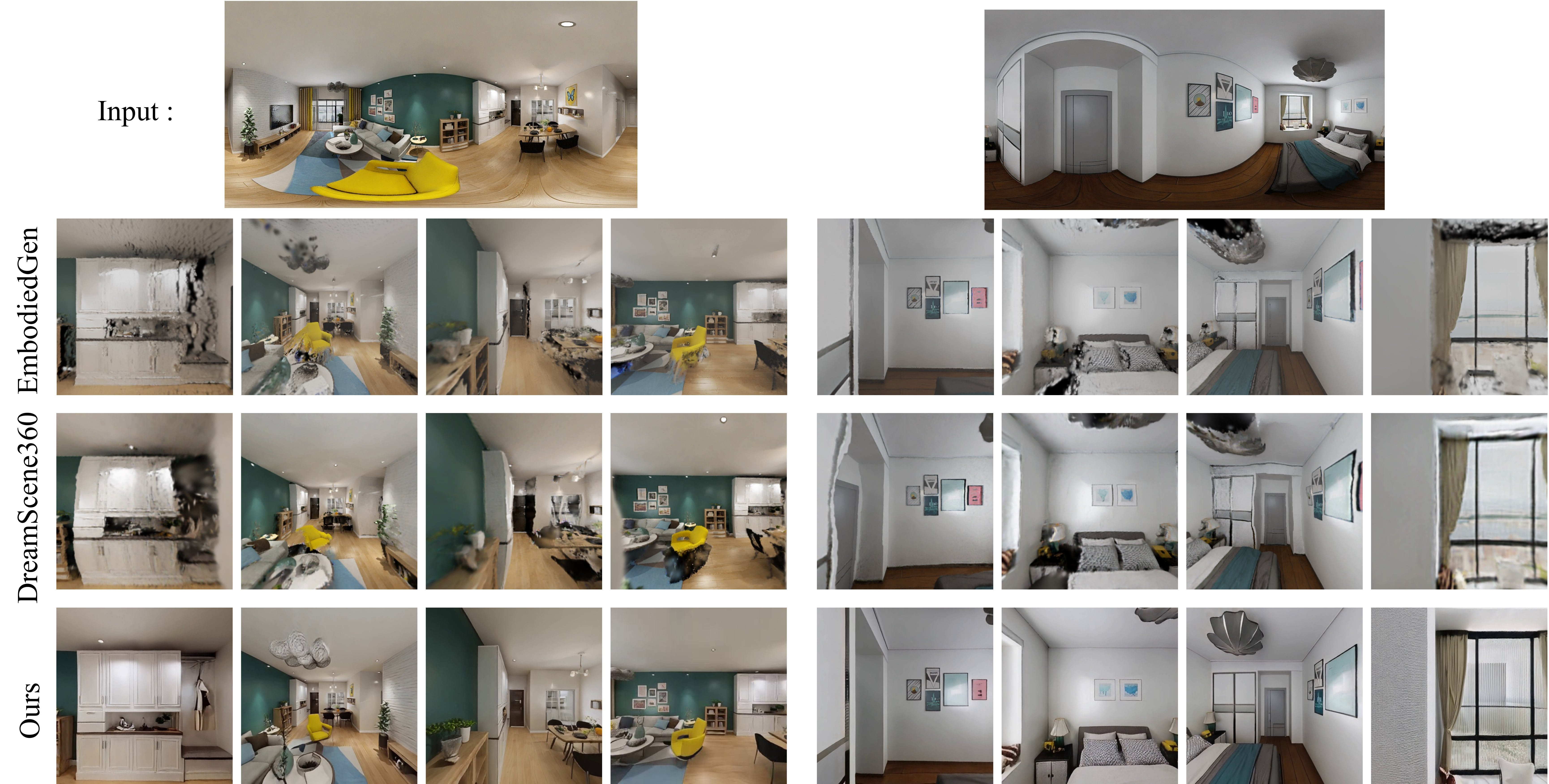}
    \caption{
    Qualitative comparison of rendered perspective views on Structured3D.
    The top row shows the input panoramas, and the following rows show novel views rendered from reconstructed 3D scenes.
    Rein3D maintains more stable geometry and fewer artifacts under large viewpoint changes.
    }
    \label{fig:img2scene_per}
\end{figure*}

We first evaluate scene reconstruction quality on 50 randomly selected scenes from the Structured3D test set~\cite{zheng2020structured3d}.
For each scene, we use a single panoramic RGB-D observation as input.
To evaluate global scene consistency under camera motion, we use the safe navigation map computed from the input depth and sample a dense candidate pool via Monte Carlo sampling.
We then apply Farthest Point Sampling (FPS) to select 5 spatially diverse camera locations, including the input panorama center and 4 distant exploratory points.
At each location, we render 4 horizontal perspective views for evaluation.

We compare Rein3D with EmbodiedGen~\cite{wang2025embodiedgen} and DreamScene360~\cite{zhou2024dreamscene360}.
Since this setting does not provide ground-truth target views for all rendered viewpoints, we report no-reference and semantic quality metrics:
Q-Align~\cite{wu2023q} for perceptual visual quality, CLIP~\cite{radford2021learning} for semantic consistency, and NIQE~\cite{mittal2012making} and BRISQUE~\cite{mittal2012no} for natural image quality.

\begin{table}[htbp]
\centering
\footnotesize
\caption{
Quantitative comparison on single-panorama 3D scene reconstruction on 50 Structured3D test scenes.
We report no-reference visual quality and semantic consistency metrics on  novel views.
}
\label{tab:structured3d_scene_quant}
\setlength{\tabcolsep}{4.5pt}
\begin{tabular}{lcccc}
\toprule
Method & Q-Align $\uparrow$ & CLIP $\uparrow$ & NIQE $\downarrow$ & BRISQUE $\downarrow$ \\
\midrule
EmbodiedGen   & 2.2914 & 0.2216 & 7.9231 & 58.3813 \\
DreamScene360 & 1.8342 & 0.2356 & 8.5295 & 61.4567 \\
\textbf{Ours} & \textbf{3.1033} & \textbf{0.2609} & \textbf{6.4925} & \textbf{46.5411} \\
\bottomrule
\end{tabular}
\end{table}

As shown in Tab.~\ref{tab:structured3d_scene_quant}, Rein3D achieves better visual quality and perceptual realism under large viewpoint changes.
Unlike methods that directly hallucinate novel views or rely on local scene expansion, our restore-and-refine pipeline first exposes missing regions through radial panoramic exploration, restores them using a panoramic video prior, and then fuses the restored observations into a global 3DGS.
This design leads to more stable rendered views and fewer geometric artifacts when the camera moves away from the input panorama center.

\subsubsection{Novel View Synthesis Evaluation on 3D-FRONT.}

To evaluate reconstruction fidelity with ground-truth supervision, we construct a novel view synthesis benchmark from 3D-FRONT.
We randomly select 30 scenes, use a single panoramic RGB-D observation as input for each scene, and evaluate all methods on the same 100 held-out target views rendered from the ground-truth 3D assets.
We compare Rein3D with EmbodiedGen under the same single-panorama input setting and report PSNR, SSIM~\cite{wang2004image}, and LPIPS.

\begin{table}[htbp]
\centering
\footnotesize
\caption{
Novel view synthesis comparison on 30 3D-FRONT scenes.
Each method takes a single panoramic RGB-D image as input and is evaluated on 100 held-out target views per scene.
}
\label{tab:3dfront_nvs}
\setlength{\tabcolsep}{6pt}
\begin{tabular}{lccc}
\toprule
Method & PSNR $\uparrow$ & SSIM $\uparrow$ & LPIPS $\downarrow$ \\
\midrule
EmbodiedGen & 15.05 & 0.517 & 0.678 \\
\textbf{Ours} & \textbf{17.64} & \textbf{0.624} & \textbf{0.540} \\
\bottomrule
\end{tabular}
\end{table}

As shown in Tab.~\ref{tab:3dfront_nvs}, Rein3D clearly outperforms EmbodiedGen across all NVS metrics.
The improvements in PSNR and SSIM indicate higher reconstruction fidelity, while the lower LPIPS demonstrates better perceptual consistency.
These results show that our restored panoramic observations provide effective pseudo supervision for refining the global 3DGS representation.

\noindent
\textbf{Qualitative comparison.}
As shown in Fig.~\ref{fig:img2scene_per}, EmbodiedGen produces floating artifacts and unstable structures under large viewpoint changes.
In contrast, Rein3D reconstructs more complete and view-consistent indoor scenes by fusing restored panoramic observations into a single global 3DGS.

\subsection{Ablation Studies}
\label{sec:ablation}

We conduct ablation studies at two levels. 
First, we analyze the key designs of the panoramic video restoration model, including the Context Anchor and spherical adaptation strategies.
Second, we evaluate how different components affect the final 3D scene reconstruction quality on 3D-FRONT.
This separation allows us to distinguish improvements in the video restoration module from improvements in the downstream 3D reconstruction pipeline.

\subsubsection{Video Restoration Ablation}

We ablate the main components of our panoramic video restoration model on the PanoV2V-15K test split. 
All variants are evaluated using the same metrics as in Tab.~\ref{tab:restoration_comparison}.
\begin{table*}[t]
\centering
\footnotesize
\caption{
Ablation study of key components in the panoramic video restoration model.
Context Anchor stabilizes long-range restoration, while latitude-aware sampling and latitude-decay loss adapt the model to equirectangular geometry.
}
\label{tab:video_component_ablation}
\setlength{\tabcolsep}{5pt}
\begin{tabular}{lccccc}
\toprule
Setting & PSNR $\uparrow$ & SSIM $\uparrow$ & WS-PSNR $\uparrow$ & WS-SSIM $\uparrow$ & FVD $\downarrow$ \\
\midrule
w/o Context Anchor & 22.84 & 0.810 & 22.40 & 0.802 & 7.75 \\
w/o Latitude-Decay Loss & 23.28 & 0.819 & 22.97 & 0.816 & 5.66 \\
w/o Latitude-Aware Noise Sampling & 22.96 & 0.816 & 22.65 & 0.812 & 6.24 \\
\textbf{Full} & \textbf{23.77} & \textbf{0.823} & \textbf{23.45} & \textbf{0.820} & \textbf{5.26} \\
\bottomrule
\end{tabular}
\end{table*}
As shown in Tab.~\ref{tab:video_component_ablation}, removing any component degrades restoration performance.
The Context Anchor is important for maintaining temporal and appearance consistency during camera exploration, while the two spherical adaptations improve reconstruction fidelity under equirectangular projection.
The full model achieves the best overall performance, confirming the effectiveness of combining global context anchoring with panorama-aware optimization.

\textbf{Sensitivity to latitude-decay coefficient.}
We further study the effect of the latitude-decay coefficient $\lambda$, which controls the strength of polar-region reweighting during training.
The setting $\lambda=1$ corresponds to a spatially uniform loss without latitude decay.

\begin{table}[t]
\centering
\footnotesize
\caption{
Sensitivity analysis of the latitude-decay coefficient $\lambda$.
The default setting $\lambda=0.1$ provides the best overall balance between spherical fidelity and temporal consistency.
}
\label{tab:lambda_sensitivity}
\setlength{\tabcolsep}{4.5pt}
\begin{tabular}{lccccc}
\toprule
Setting & PSNR $\uparrow$ & SSIM $\uparrow$ & WS-PSNR $\uparrow$ & WS-SSIM $\uparrow$ & FVD $\downarrow$ \\
\midrule
$\lambda=1.0$ & 23.28 & 0.819 & 22.97 & 0.816 & 5.66 \\
$\lambda=0.5$ & \textbf{23.80} & 0.822 & 23.38 & 0.818 & 5.66 \\
$\lambda=0.3$ & 23.42 & 0.819 & 23.05 & 0.816 & 5.45 \\
$\lambda=0$ & 23.44 & \textbf{0.824} & 22.99 & 0.819 & 5.62 \\
$\lambda=0.1$ \textbf{(Full)} & 23.77 & 0.823 & \textbf{23.45} & \textbf{0.820} & \textbf{5.26} \\
\bottomrule
\end{tabular}
\end{table}

Tab.~\ref{tab:lambda_sensitivity} shows that the model is sensitive to the latitude-decay coefficient.
Weak or strong polar reweighting leads to suboptimal trade-offs between standard image quality, spherical fidelity, and temporal consistency.
We use $\lambda=0.1$ as the default setting in all experiments.
\subsubsection{3D Scene Reconstruction Ablation}

We evaluate the impact of key pipeline components on the 3D-FRONT NVS benchmark, where ground-truth target views are available.
All variants take a single panoramic RGB-D image as input.

\begin{table}[htbp]
\centering
\footnotesize
\caption{
Ablation study on 3D-FRONT.
w/o Radial Exploration uses local exploration around the input panorama center.
w/o VSR removes super-resolution before 3DGS refinement.
$\tau$ denotes the number of radial trajectories.
}
\label{tab:scene_ablation}
\setlength{\tabcolsep}{6pt}
\begin{tabular}{lccc}
\toprule
Setting & PSNR $\uparrow$ & SSIM $\uparrow$ & LPIPS $\downarrow$ \\
\midrule
w/o Radial Exploration & 14.84 & 0.448 & 0.672 \\
w/o VSR & 17.42 & 0.601 & 0.549 \\
Full ($\tau=4$) & 17.36 & 0.595 & 0.553 \\
\textbf{Full ($\tau=8$)} & \textbf{17.64} & \textbf{0.624} & \textbf{0.540} \\
\bottomrule
\end{tabular}
\end{table}

As shown in Tab.~\ref{tab:scene_ablation}, radial exploration is critical for exposing disoccluded regions under large camera motion.
Removing VSR slightly degrades reconstruction quality, indicating that high-resolution pseudo observations benefit final 3DGS refinement.
Increasing the number of radial trajectories from $\tau=4$ to $\tau=8$ further improves reconstruction quality by providing broader scene coverage.
We therefore use $\tau=8$ as the default setting.

\section{Conclusion}
We presented Rein3D, a restore-and-refine framework for single-panorama 3D indoor scene reconstruction.
Rein3D lifts an input panoramic RGB-D observation into a coarse 3DGS, restores degraded RGB-alpha panoramic videos rendered along radial trajectories, and fuses the restored pseudo observations back into a global 3DGS.
We also introduced PanoV2V-15K, a paired clean--degraded panoramic video dataset for learning this restoration task.
Experiments show that Rein3D improves video restoration fidelity, view consistency, and long-range novel-view rendering over existing baselines.









\bibliographystyle{unsrtnat}
\bibliography{main}

\newpage

\appendix
\section{Detailed Overview of PanoV2V-15K Dataset}

In this section, we introduced PanoV2V-15K, a large-scale dataset comprising over 15,050 paired clean and degraded panoramic videos tailored for diffusion-based scene restoration. This section provides a more comprehensive breakdown of the dataset's composition, the diversity of the indoor environments, and the pipeline used to acquire high-fidelity priors.

\begin{figure*}[htbp] 
    \centering
    \includegraphics[width=\linewidth]{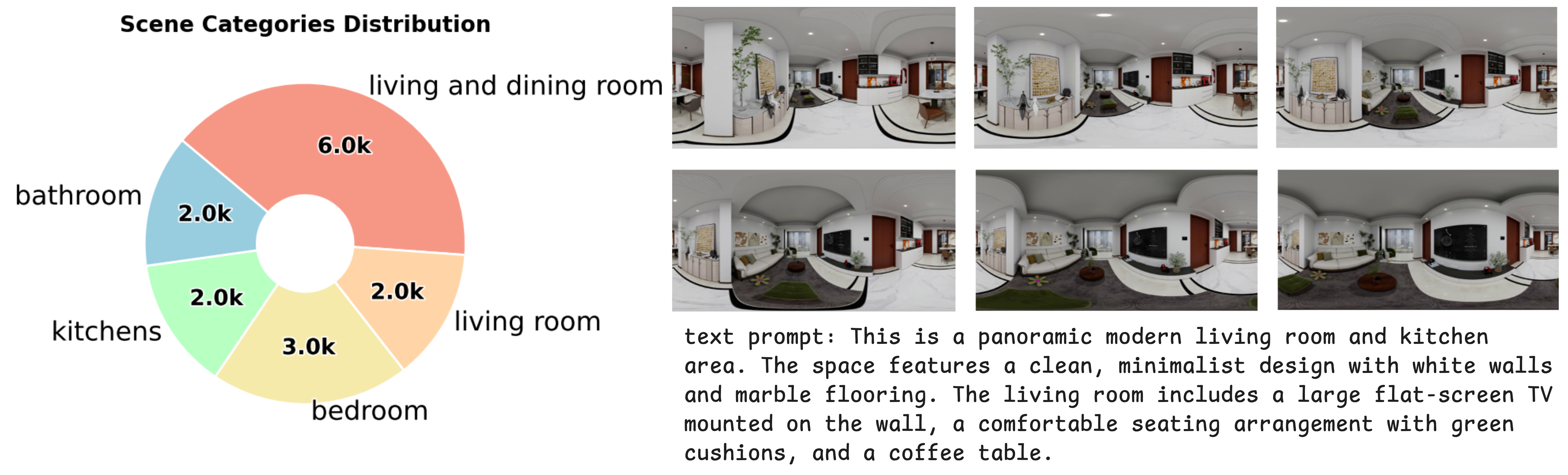}
    \caption{Overview of the PanoV2V-15K dataset. \textbf{(Left)} A pie chart illustrating the distribution of the 15,500 indoor scenes across different room types. \textbf{(Right)} Consecutive frames extracted from a high-quality panoramic video within the dataset, accompanied by its corresponding highly descriptive text prompt generated by Qwen2.5-VL.}
    \label{fig:dataset_composition}
    
    \vspace{0.5cm} 
    
    \centering
    \includegraphics[width=\linewidth]{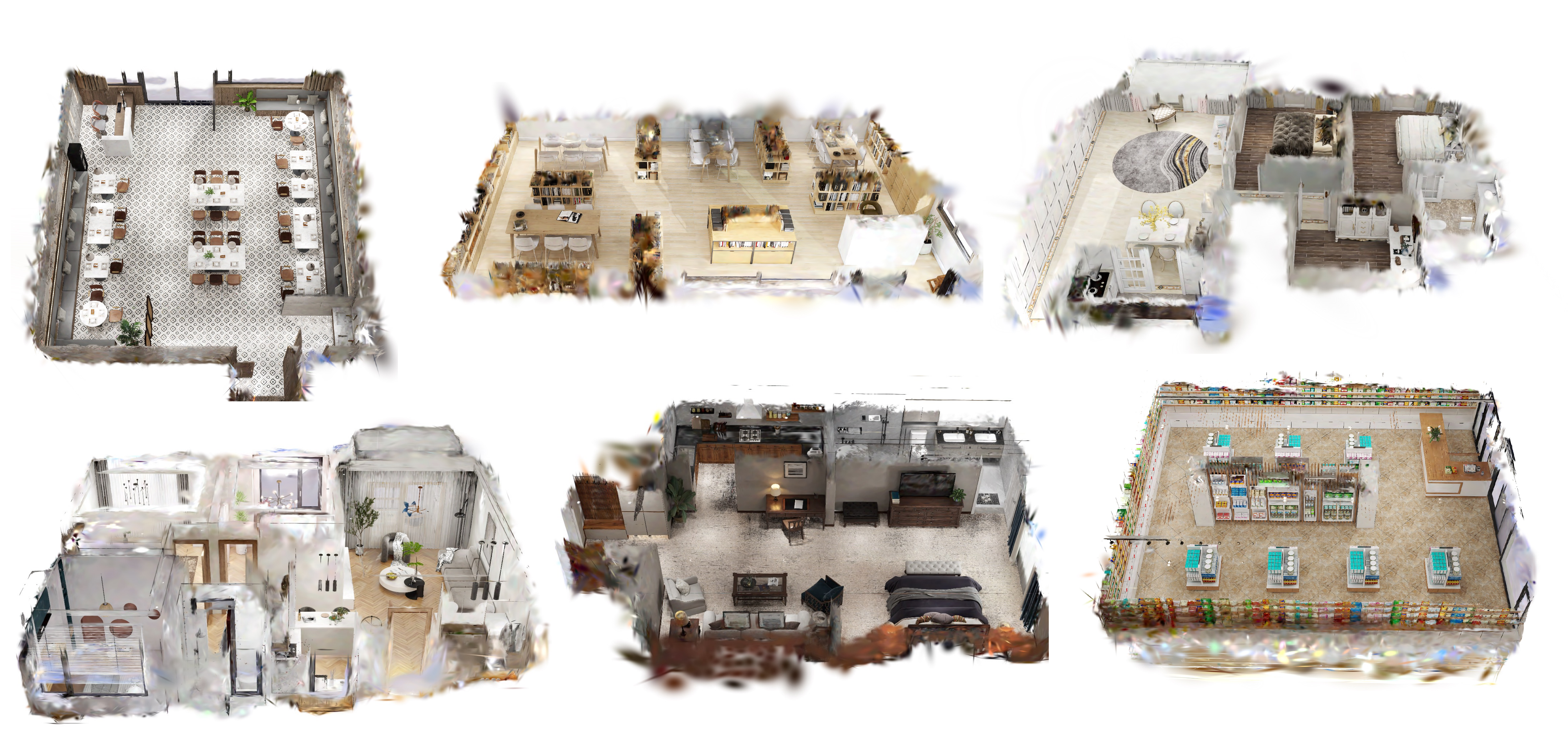}
    \caption{Top-down visualizations of six representative scenes from the PanoV2V-15K dataset. The original 3D mesh environments are successfully reconstructed and visualized by 3D Gaussian Splatting, demonstrating high-fidelity geometry, coherent global layouts, and rich textural details across diverse room categories.}
    \label{fig:3dgs_topdown}
\end{figure*}

To ensure broad scene diversity and generalizability for the generation model, the PanoV2V-15K dataset consists of 15,050 distinct, professionally designed indoor environments. As illustrated in the pie chart in Fig. \ref{fig:dataset_composition} (Left), the dataset covers a wide variety of functional layouts, including 6,050 living and dining rooms, 2,000 separate living rooms, 3,000 bedrooms, 2,000 kitchens, and 2,000 bathrooms. This diverse distribution ensures that our model learns a robust and comprehensive structural prior for complex indoor configurations.

Originally, these high-quality environments were represented as dense 3D meshes. To facilitate efficient and photo-realistic rendering for our panoramic video generation pipeline, all scenes were meticulously reconstructed and are visualized by 3D Gaussian Splatting (3DGS)\cite{kerbl20233d}. As demonstrated by the top-down views in Fig. \ref{fig:3dgs_topdown}, this conversion from mesh to 3DGS preserves fine-grained geometric details and complex textures, providing a reliable foundation for our panoramic renderings. 

We additionally utilized the advanced vision-language model, Qwen2.5-VL\cite{bai2025qwen3}, to automatically generate highly descriptive and semantically rich text prompts for each scene based on the rendered panoramic views. An example of consecutive panoramic video frames paired with their corresponding generated text prompt is provided in Fig. \ref{fig:dataset_composition} (Right).

\vspace{1cm}

\section{3DGS Refinement Details}
\label{app:3dgs_refinement}

We refine the final 3DGS representation using \texttt{refine\_3dgs.py}. 
The main training hyperparameters are summarized in Tab.~\ref{tab:3dgs_refine_details}.

\begin{table}[h]
\centering
\caption{Key hyperparameters for final 3DGS refinement.}
\label{tab:3dgs_refine_details}
\setlength{\tabcolsep}{6pt}
\begin{tabular}{lc}
\toprule
Setting & Value \\
\midrule
Batch size & 4 \\
Training steps & 15,000 \\
Evaluation / save steps & 5,000 / 10,000 / 15,000 \\
Test split & Every 16th view \\
SH degree & 3 \\
Near / far plane & 0.01 / 100.0 \\
Anti-aliasing & Enabled \\
SSIM loss weight & 0.2 \\
\midrule
Densification start / stop & 500 / 13,000 \\
Densification interval & 100 \\
Opacity pruning threshold & 0.005 \\
Opacity reset interval & 3,000 \\
\midrule
LR for means & $1.6\times10^{-4}$ \\
LR for scales & $5.0\times10^{-3}$ \\
LR for opacities & $5.0\times10^{-2}$ \\
LR for rotations & $1.0\times10^{-3}$ \\
LR for SH coefficients & $2.5\times10^{-3}$ / $1.25\times10^{-4}$ \\
\bottomrule
\end{tabular}
\end{table}

\section{Algorithmic Details for Camera Exploration and Evaluation}

\begin{figure}[htbp]
    \centering
    \includegraphics[width=\linewidth]{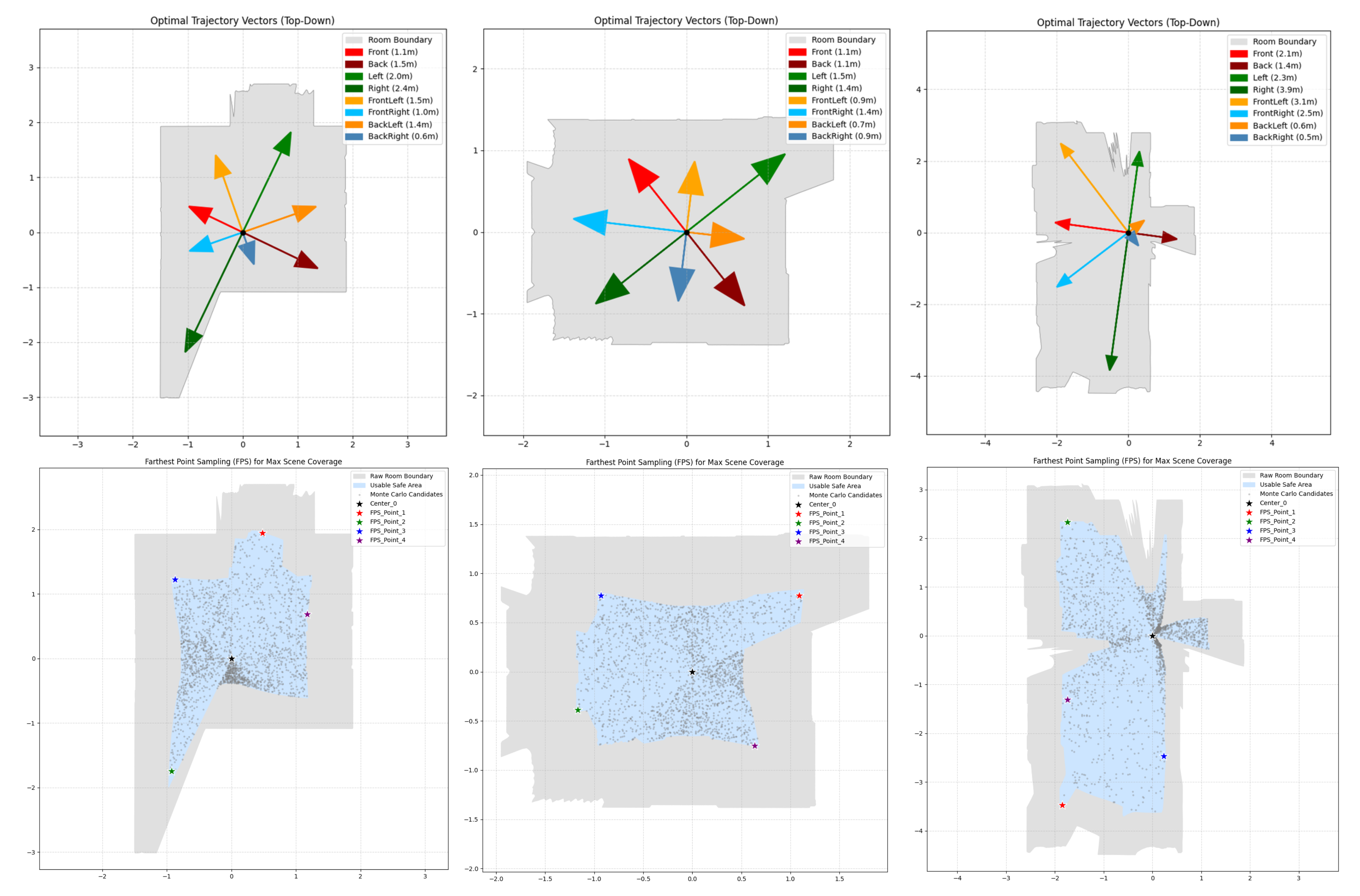}
    \caption{Visualizations of our camera exploration and evaluation sampling strategies. \textbf{(Top)} The depth-optimized radial trajectory generation, illustrating the safe profile, the cross-score optimization, and the selected top-down trajectories. \textbf{(Bottom)} The Farthest Point Sampling (FPS) process for evaluation, showing the safe navigable area, the Monte Carlo candidate pool, and the selected 5 globally distributed camera locations.}
    \label{fig:sampling_vis}
\end{figure}

In this section, we briefly introduced our trajectory sampling strategy for scene exploration and the Farthest Point Sampling (FPS) strategy for objective evaluation. To provide a more intuitive understanding, we visualize these mechanisms in Figure~\ref{fig:sampling_vis}, which illustrates the depth-optimized radial trajectory generation (top) alongside the global view sampling via FPS (bottom). To further facilitate reproducibility, we provide the detailed pseudocodes for both processes in this section.

\subsection{Radial Trajectory Generation via Depth Optimization}
To effectively uncover missing regions while preventing camera collisions, we employ a robust radial trajectory sampling strategy tailored for panoramic depth maps. Rather than computing a dense 2D safe navigation map, we optimize the trajectories directly through 1D depth profile analysis, which is highly efficient.

Specifically, we first extract a central horizontal band from the estimated panoramic depth (e.g., spanning $45\%$ to $55\%$ of the image height) and compress it along the vertical axis by taking the minimum depth. This yields a raw 1D depth profile that accounts for structural obstacles at the camera's height. To ensure physical safety during movement, we apply a 1D minimum filter with a wrap mode to this profile, essentially performing a morphological erosion that accommodates the camera's physical radius ($r_c$) and a predefined safety buffer ($b_s$). 

To maximize the total observable area, we define $\tau$ uniformly distributed directional vectors radiating from the origin (e.g., $\tau=4$ for a cross-shaped orthogonal exploration). We dynamically search for an optimal global rotation offset $\theta^*$ that maximizes the sum of the safe travel distances along all $\tau$ directions. Finally, the trajectories are sampled along these optimized vectors, incorporating a Y-axis inversion to correctly align the 2D equirectangular projection with the 3D world coordinate system. The detailed procedure is outlined in Algorithm \ref{alg:radial_search}.

\begin{algorithm}[htbp]
\caption{Radial Trajectory Generation via Depth Optimization}
\label{alg:radial_search}
\begin{algorithmic}[1]
\Require Panoramic depth map $\mathcal{D}$, Number of trajectories $\tau$ (e.g., 4), Camera radius $r_c$, Stop buffer $b_s$, Step size $\Delta d$.
\Ensure Set of $\tau$ radial trajectories $\mathcal{T}$.
\State Extract the central horizontal depth band from $\mathcal{D}$ (e.g., height ratio $[0.45, 0.55]$).
\State Compute raw depth profile $D_{raw}$ by taking the minimum depth along the vertical axis of the extracted band.
\State Apply a 1D minimum filter (with wrap mode) to $D_{raw}$ to obtain the eroded safe profile $D_{safe}$.
\State Initialize maximum cross-score $S_{max} = 0$, optimal angle offset $\theta^* = 0$.
\For{offset $\theta \in [0, \frac{2\pi}{\tau})$}
    \State Compute trajectory score: $S_{curr} = \sum_{i=0}^{\tau-1} D_{safe}(\theta + i \frac{2\pi}{\tau})$.
    \If{$S_{curr} > S_{max}$}
        \State $S_{max} = S_{curr}$
        \State $\theta^* = \theta$
    \EndIf
\EndFor
\State Initialize trajectory set $\mathcal{T} = \emptyset$.
\For{$i = 0$ to $\tau-1$}
    \State Target angle $\phi = \theta^* + i \frac{2\pi}{\tau}$.
    \State Fetch safe distance for this direction: $d_{safe} = D_{safe}(\phi)$.
    \State Calculate maximum travel distance: $l = \max(0, d_{safe} - r_c - b_s)$.
    \State Compute 3D direction vector: $\mathbf{v} = (\cos\phi, -\sin\phi, 0)$ \Comment{Inverted Y-axis alignment}
    \State Sample trajectory points: $P = \{ k \cdot \Delta d \cdot \mathbf{v} \mid k \in \mathbb{N}, k \cdot \Delta d \le l \}$.
    \State $\mathcal{T} = \mathcal{T} \cup \{ P \}$.
\EndFor
\State \Return $\mathcal{T}$
\end{algorithmic}
\end{algorithm}

\subsection{Farthest Point Sampling for Evaluation Viewpoints}

To ensure a robust and fair quantitative assessment, we must evaluate the generated scenes from viewpoints far from the initialization point. Directly sampling on a dense grid can be inefficient and heavily biased toward the scene center. Instead, we utilize the safe depth profile obtained during the radial trajectory generation to establish a strictly usable navigation area and apply Farthest Point Sampling (FPS) to deterministically select 5 spatially diverse camera locations. 

Specifically, we first calculate a usable depth profile by further shrinking the safe profile by the camera radius and a safety buffer, masking out extremely narrow gaps to prevent clipping artifacts. We then generate a dense candidate pool via area-uniform Monte Carlo sampling. To ensure uniform spatial distribution within the radial sectors, the sampled distance is scaled by the square root of a uniform random variable. Farthest Point Sampling is subsequently applied, strictly initializing the origin as the first selected point, ensuring that the subsequent selections are pushed as close to the valid boundaries as possible. At each of the 5 locations, 4 horizontal perspective views are rendered to ensure comprehensive visual coverage. The complete process is detailed in Algorithm \ref{alg:fps_sampling}.

\begin{algorithm}[htbp]
\caption{Area-Uniform Monte Carlo and Farthest Point Sampling}
\label{alg:fps_sampling}
\begin{algorithmic}[1]
\Require Safe depth profile $D_{safe}$, Number of additional FPS points $N_{fps}$ (e.g., 4), Camera radius $r_c$, Stop buffer $b_s$, Number of Monte Carlo samples $M$ (e.g., 2000).
\Ensure Set of evaluation camera poses $\mathcal{P}$.
\State Compute strictly usable depth profile: $D_{usable} = \max(0, D_{safe} - r_c - b_s)$.
\State Initialize candidate pool $\mathcal{C} = \emptyset$.
\For{$m = 1$ to $M$}
    \State Sample a random angle index $idx \in [0, |D_{usable}|-1]$ and get $\theta = \text{angle\_at}(idx)$.
    \State Fetch maximum usable radius $R_{max} = D_{usable}[idx]$.
    \If{$R_{max} > 0.1$} \Comment{Ignore extremely narrow gaps}
        \State Area-uniform radius sampling: $r = R_{max} \cdot \sqrt{\mathrm{rand}(0, 1)}$.
        \State Convert to 3D coordinates: $\mathbf{c} = (r \cos\theta, -r \sin\theta, 0)$. \Comment{Inverted Y-axis}
        \State $\mathcal{C} = \mathcal{C} \cup \{\mathbf{c}\}$.
    \EndIf
\EndFor
\State Initialize selected locations $V = \{\mathbf{0}\}$. \Comment{Force origin as the first point}
\While{$|V| < N_{fps} + 1$}
    \State Initialize maximum distance $d_{max} = -1$, best candidate $\mathbf{c}^* = \mathrm{null}$.
    \For{each candidate $\mathbf{c} \in \mathcal{C} \setminus V$}
        \State Compute shortest distance to already selected points: $d_{min} = \min_{\mathbf{v} \in V} \|\mathbf{c} - \mathbf{v}\|_2$.
        \If{$d_{min} > d_{max}$}
            \State $d_{max} = d_{min}$
            \State $\mathbf{c}^* = \mathbf{c}$
        \EndIf
    \EndFor
    \State $V = V \cup \{\mathbf{c}^*\}$.
\EndWhile
\State Initialize pose set $\mathcal{P} = \emptyset$.
\For{each location $\mathbf{v} \in V$}
    \For{yaw angle $\psi \in \{0, \frac{\pi}{2}, \pi, \frac{3\pi}{2}\}$}
        \State Add camera pose $(\mathbf{v}, \psi)$ to $\mathcal{P}$.
    \EndFor
\EndFor
\State \Return $\mathcal{P}$
\end{algorithmic}
\end{algorithm}

\section{Additional Qualitative Results}

In this section, we provide further qualitative results to demonstrate the effectiveness of our \textit{restore-and-refine} paradigm. We first showcase the intermediate panoramic video restoration results, followed by comprehensive comparisons of the final generated 3D scenes against state-of-the-art baselines.

\subsection{Panoramic Video Restoration}
\begin{figure*}[htbp]
    \centering
    \includegraphics[width=\linewidth]{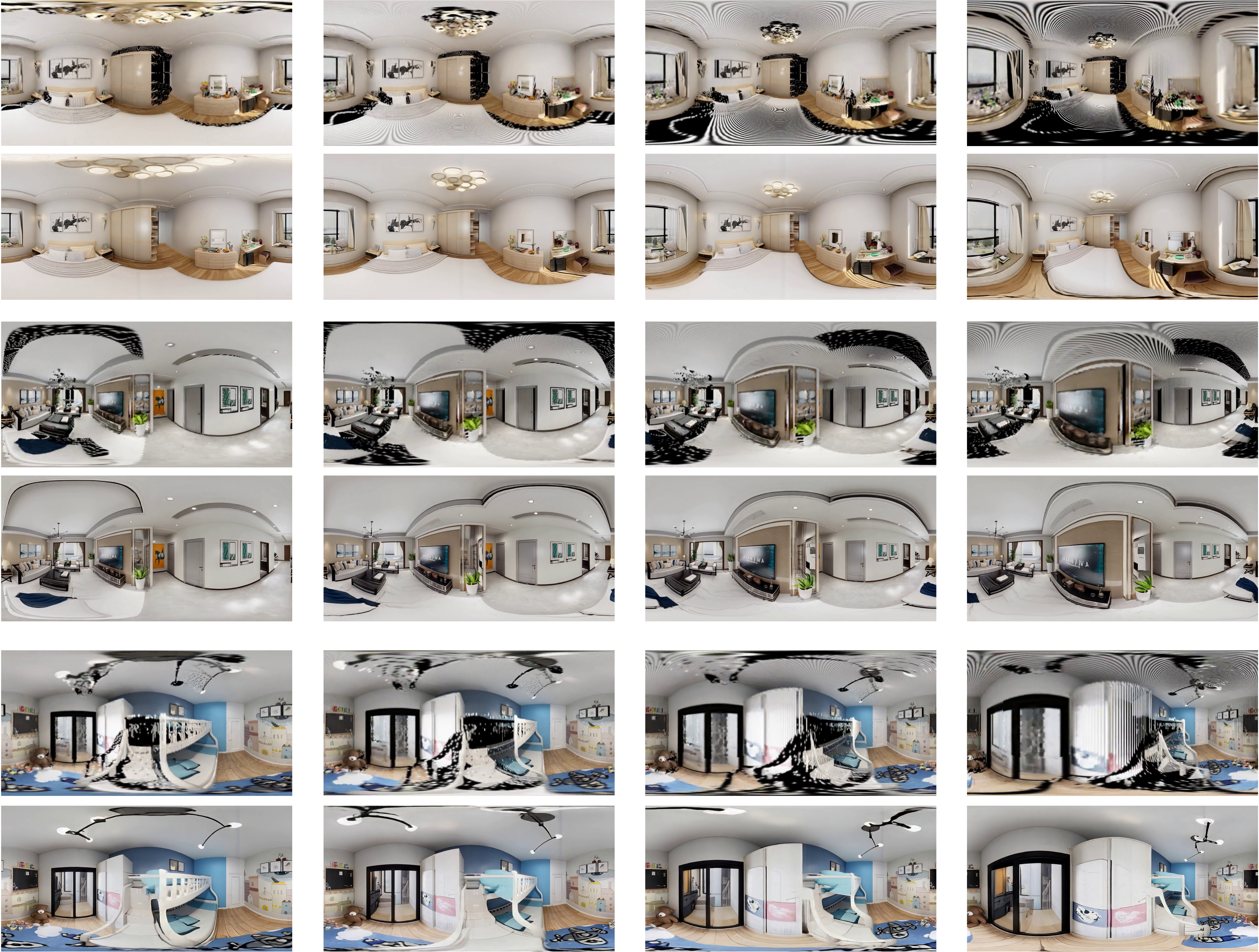}
    \caption{Qualitative results of panoramic video restoration. The odd-numbered rows (1st, 3rd, and 5th) show sequences of degraded panoramic frames rendered from the initial coarse 3DGS, which suffer from severe missing regions (black holes) and structural artifacts. Directly below them, the even-numbered rows (2nd, 4th, and 6th) display the corresponding high-fidelity frames restored by our panoramic video diffusion model, demonstrating remarkable spatial completeness and temporal coherence across the trajectory.}
    \label{fig:supp_video_restoration}
\end{figure*}

As discussed in the main paper, rendering a coarse 3D Gaussian Splatting (3DGS) scene along radial trajectories naturally yields degraded panoramic sequences due to massive missing geometry in unseen areas. In Fig. \ref{fig:supp_video_restoration}, we visualize this restoration process. Given the highly degraded and incomplete RGB-Alpha panoramic inputs, our panoramic video diffusion model effectively hallucinates the missing regions while maintaining strict temporal and spatial consistency. The restored frames exhibit photorealistic textures and coherent structures, successfully eliminating the transparent holes and visual artifacts present in the raw renderings.

\subsection{3D Scene Generation and Novel View Synthesis}
To evaluate the global consistency and structural integrity of the final reconstructed 3D environments, we compare Rein3D against two recent unconstrained 3D scene generation baselines: EmbodiedGen\cite{wang2025embodiedgen} and DreamScene360\cite{zhou2024dreamscene360}. 

\begin{figure*}[htbp]
    \centering
    \includegraphics[width=\linewidth]{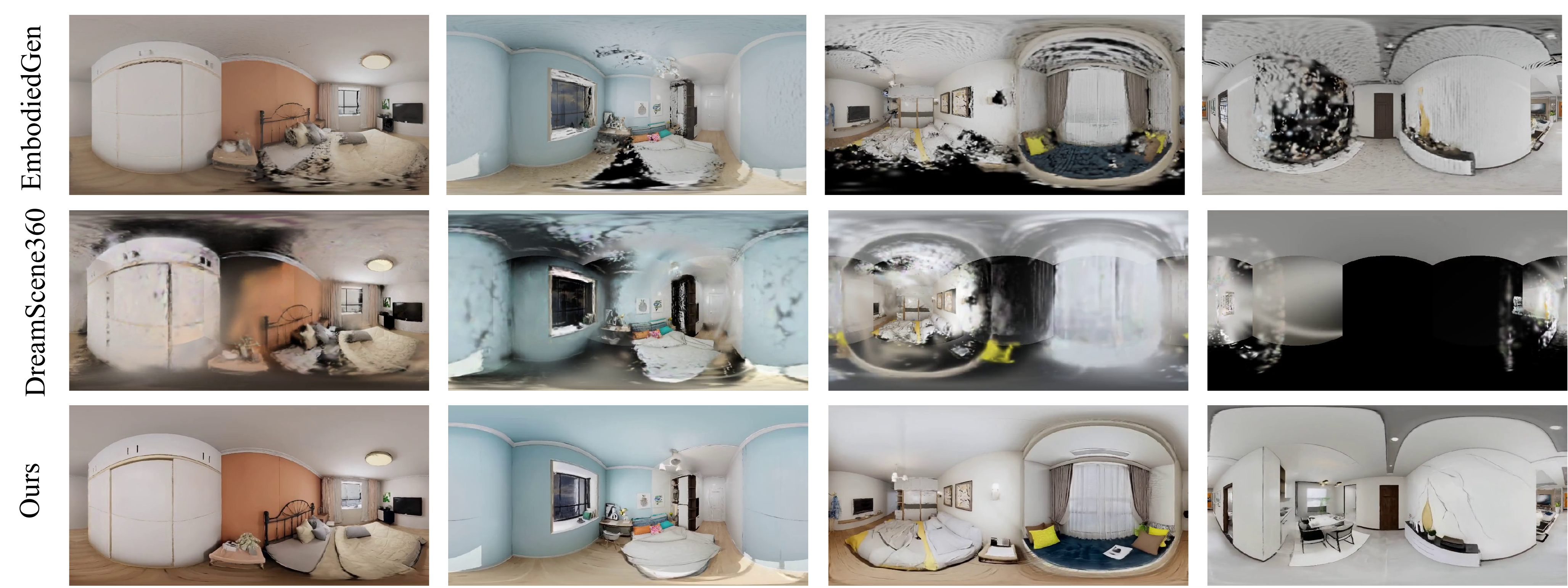}
    \caption{Additional qualitative comparisons on novel view synthesis against EmbodiedGen and DreamScene360. The views are rendered from camera locations far from the initialization point to evaluate long-range exploration consistency. While baselines exhibit severe distortions, floating artifacts, and missing structures, our method preserves global geometry and fine textural details.}
    \label{fig:supp_baseline_comparison}
\end{figure*}

As shown in Fig. \ref{fig:supp_baseline_comparison}, when the camera translates significantly away from the starting anchor, existing methods struggle to maintain global coherence. EmbodiedGen, which relies heavily on localized image inpainting, tends to generate unstructured floating artifacts resembling messy point clouds in unobserved regions. DreamScene360 frequently suffers from severe global geometric distortion, warping the scene structure and introducing conspicuous voids. In contrast, our Rein3D framework, guided by temporally coherent panoramic video priors, generates stable, photorealistic, and geometrically consistent novel views even under large camera displacements.

\newpage

\section*{NeurIPS Paper Checklist}

\begin{enumerate}

\item {\bf Claims}
    \item[] Question: Do the main claims made in the abstract and introduction accurately reflect the paper's contributions and scope?
    \item[] Answer: \answerYes{}.
    \item[] Justification: The abstract and introduction state that Rein3D reconstructs complete and explorable 3D indoor scenes from a single panoramic RGB-D observation by formulating scene completion as 3D-prior-guided panoramic video restoration followed by global 3DGS refinement. The main claims are supported by panoramic video restoration experiments on PanoV2V-15K and 3D-FRONT, 3D scene reconstruction experiments on Structured3D and 3D-FRONT, and ablation studies of the key components.
    \item[] Guidelines:
    \begin{itemize}
        \item The answer \answerNA{} means that the abstract and introduction do not include the claims made in the paper.
        \item The abstract and/or introduction should clearly state the claims made, including the contributions made in the paper and important assumptions and limitations.
        \item The claims made should match theoretical and experimental results, and reflect how much the results can be expected to generalize to other settings.
        \item It is fine to include aspirational goals as motivation as long as it is clear that these goals are not attained by the paper.
    \end{itemize}

\item {\bf Limitations}
    \item[] Question: Does the paper discuss the limitations of the work performed by the authors?
    \item[] Answer: \answerYes{}.
    \item[] Justification: The paper discusses limitations related to the dependence on input depth quality, the focus on static indoor scenes, possible inconsistencies introduced by diffusion-based restoration, and the computational cost of video restoration and 3DGS refinement. We also note that further validation on real captured panoramas and more complex dynamic scenes is an important direction for future work.
    \item[] Guidelines:
    \begin{itemize}
        \item The answer \answerNA{} means that the paper has no limitation while the answer \answerNo{} means that the paper has limitations, but those are not discussed in the paper.
        \item The authors are encouraged to create a separate ``Limitations'' section in their paper.
        \item The paper should point out any strong assumptions and how robust the results are to violations of these assumptions.
        \item The authors should reflect on the scope of the claims made.
        \item The authors should discuss the computational efficiency of the proposed algorithms and how they scale with dataset size.
    \end{itemize}

\item {\bf Theory assumptions and proofs}
    \item[] Question: For each theoretical result, does the paper provide the full set of assumptions and a complete proof?
    \item[] Answer: \answerNA{}.
    \item[] Justification: The paper does not present formal theoretical results, theorems, or proofs. The proposed approach is an empirical framework for panoramic video restoration and 3DGS-based single-panorama indoor scene reconstruction.
    \item[] Guidelines:
    \begin{itemize}
        \item The answer \answerNA{} means that the paper does not include theoretical results.
        \item All the theorems, formulas, and proofs in the paper should be numbered and cross-referenced.
        \item All assumptions should be clearly stated or referenced in the statement of any theorems.
        \item The proofs can either appear in the main paper or the supplemental material.
    \end{itemize}

\item {\bf Experimental result reproducibility}
    \item[] Question: Does the paper fully disclose all the information needed to reproduce the main experimental results?
    \item[] Answer: \answerYes{}.
    \item[] Justification: The paper describes the PanoV2V-15K construction protocol, coarse 3DGS initialization, radial camera exploration, panoramic video restoration network, training settings, evaluation datasets, metrics, and ablation protocols. Additional details for camera sampling, evaluation viewpoint selection, and 3DGS refinement hyperparameters are provided in the appendix.
    \item[] Guidelines:
    \begin{itemize}
        \item The answer \answerNA{} means that the paper does not include experiments.
        \item If the paper includes experiments, a \answerNo{} answer to this question will not be perceived well by the reviewers.
        \item If the contribution is a dataset and\slash or model, the authors should describe the steps taken to make their results reproducible or verifiable.
        \item Reproducibility can be accomplished by releasing code and data, providing detailed instructions, releasing model checkpoints, or other appropriate means.
    \end{itemize}

\item {\bf Open access to data and code}
    \item[] Question: Does the paper provide open access to the data and code, with sufficient instructions to faithfully reproduce the main experimental results, as described in supplemental material?
    \item[] Answer: \answerYes{}.
    \item[] Justification: We provide anonymized access to the code, dataset construction scripts, evaluation scripts, trained model checkpoints, and documentation for PanoV2V-15K. The release includes instructions for preparing the clean--degraded panoramic video pairs, training the panoramic restoration model, running 3DGS refinement, and reproducing the main video restoration and 3D scene reconstruction experiments. Dataset access and licensing information are documented with the released assets.
    \item[] Guidelines:
    \begin{itemize}
        \item The answer \answerNA{} means that paper does not include experiments requiring code.
        \item Please see the NeurIPS code and data submission guidelines for more details.
        \item The instructions should contain the exact command and environment needed to reproduce the results.
        \item The authors should provide instructions on data access and preparation, including raw data, preprocessed data, intermediate data, and generated data.
        \item The authors should provide scripts to reproduce all experimental results for the proposed method and baselines, or state which experiments are omitted and why.
        \item At submission time, to preserve anonymity, the authors should release anonymized versions if applicable.
    \end{itemize}

\item {\bf Experimental setting/details}
    \item[] Question: Does the paper specify all the training and test details necessary to understand the results?
    \item[] Answer: \answerYes{}.
    \item[] Justification: The paper specifies the video restoration training setup, optimizer, learning rate, input resolution, clip length, batch size, training hardware, 3DGS refinement steps, anti-aliasing setting, evaluation splits, metrics, and camera sampling protocols. Additional hyperparameters for final 3DGS refinement are provided in the appendix.
    \item[] Guidelines:
    \begin{itemize}
        \item The answer \answerNA{} means that the paper does not include experiments.
        \item The experimental setting should be presented in the core of the paper to a level of detail necessary to appreciate the results and make sense of them.
        \item The full details can be provided either with the code, in appendix, or as supplemental material.
    \end{itemize}

\item {\bf Experiment statistical significance}
    \item[] Question: Does the paper report error bars suitably and correctly defined or other appropriate information about statistical significance?
    \item[] Answer: \answerNo{}.
    \item[] Justification: We report average metrics over multiple held-out scenes and video clips, including 50 PanoV2V-15K video clips, 50 3D-FRONT video restoration clips, 50 Structured3D scenes, and 30 3D-FRONT scenes with 100 target views per scene. We do not currently report error bars or confidence intervals. However, all methods are evaluated on the same fixed test sets under identical protocols.
    \item[] Guidelines:
    \begin{itemize}
        \item The answer \answerNA{} means that the paper does not include experiments.
        \item The authors should answer \answerYes{} if the results are accompanied by error bars, confidence intervals, or statistical significance tests.
        \item The factors of variability that the error bars are capturing should be clearly stated.
        \item The method for calculating the error bars should be explained.
        \item It should be clear whether the error bar is the standard deviation or the standard error of the mean.
    \end{itemize}

\item {\bf Experiments compute resources}
    \item[] Question: For each experiment, does the paper provide sufficient information on the computer resources needed to reproduce the experiments?
    \item[] Answer: \answerYes{}.
    \item[] Justification: The implementation details specify the hardware used for training the panoramic video restoration model, including 4 NVIDIA H200 GPUs, input resolution, batch size, and training steps. The appendix reports the main hyperparameters for final 3DGS refinement. We also describe the evaluation protocols for video restoration and 3D reconstruction.
    \item[] Guidelines:
    \begin{itemize}
        \item The answer \answerNA{} means that the paper does not include experiments.
        \item The paper should indicate the type of compute workers, CPU or GPU, internal cluster or cloud provider, including relevant memory and storage.
        \item The paper should provide the amount of compute required for each individual experimental run and estimate the total compute.
        \item The paper should disclose whether the full research project required more compute than the experiments reported in the paper.
    \end{itemize}

\item {\bf Code of ethics}
    \item[] Question: Does the research conducted in the paper conform with the NeurIPS Code of Ethics?
    \item[] Answer: \answerYes{}.
    \item[] Justification: The work uses synthetic indoor scenes and public research datasets for panoramic video restoration and 3D scene reconstruction. It does not involve privacy-sensitive personal data, human-subject experiments, deceptive data collection, or safety-critical deployment.
    \item[] Guidelines:
    \begin{itemize}
        \item The answer \answerNA{} means that the authors have not reviewed the NeurIPS Code of Ethics.
        \item If the authors answer \answerNo{}, they should explain the special circumstances that require a deviation from the Code of Ethics.
        \item The authors should make sure to preserve anonymity.
    \end{itemize}

\item {\bf Broader impacts}
    \item[] Question: Does the paper discuss both potential positive societal impacts and negative societal impacts of the work performed?
    \item[] Answer: \answerYes{}.
    \item[] Justification: The paper discusses positive applications in VR/AR content creation, simulation, and Embodied AI. It also notes possible risks and limitations, such as misuse of generated 3D environments, over-reliance on synthetic reconstructions, and artifacts when the method is applied outside the intended static indoor-scene setting.
    \item[] Guidelines:
    \begin{itemize}
        \item The answer \answerNA{} means that there is no societal impact of the work performed.
        \item If the authors answer \answerNA{} or \answerNo{}, they should explain why their work has no societal impact or why the paper does not address societal impact.
        \item The authors should consider possible harms that could arise when the technology is used as intended, when it gives incorrect results, and following intentional or unintentional misuse.
        \item If there are negative societal impacts, the authors could also discuss possible mitigation strategies.
    \end{itemize}

\item {\bf Safeguards}
    \item[] Question: Does the paper describe safeguards that have been put in place for responsible release of data or models that have a high risk for misuse?
    \item[] Answer: \answerNA{}.
    \item[] Justification: The work does not release a high-risk pretrained language model, human-image generator, scraped dataset, or model involving personal data. The proposed dataset consists of synthetic indoor scenes and paired clean--degraded panoramic videos for 3D reconstruction research. The released restoration model is specialized for indoor RGB-alpha panoramic restoration rather than general-purpose image or video generation.
    \item[] Guidelines:
    \begin{itemize}
        \item The answer \answerNA{} means that the paper poses no such risks.
        \item Released models that have a high risk for misuse or dual-use should be released with necessary safeguards.
        \item Datasets that have been scraped from the Internet could pose safety risks. The authors should describe how they avoided releasing unsafe images.
        \item We recognize that providing effective safeguards is challenging, but encourage authors to take this into account.
    \end{itemize}

\item {\bf Licenses for existing assets}
    \item[] Question: Are the creators or original owners of assets used in the paper properly credited and are the license and terms of use explicitly mentioned and properly respected?
    \item[] Answer: \answerYes{}.
    \item[] Justification: We cite and credit the datasets, models, and codebases used in the paper, including Structured3D, 3D-FRONT, VACE/Wan2.1, ProPainter, FlashVSR, gsplat, and 3D Gaussian Splatting. We use these assets for research purposes and follow their intended usage and license terms. The derived PanoV2V-15K dataset is released under terms compatible with the underlying indoor assets.
    \item[] Guidelines:
    \begin{itemize}
        \item The answer \answerNA{} means that the paper does not use existing assets.
        \item The authors should cite the original paper that produced the code package or dataset.
        \item The authors should state which version of the asset is used and, if possible, include a URL.
        \item The name of the license should be included for each asset.
        \item For scraped data from a particular source, the copyright and terms of service of that source should be provided.
        \item For existing datasets that are re-packaged, both the original license and the license of the derived asset should be provided.
    \end{itemize}

\item {\bf New assets}
    \item[] Question: Are new assets introduced in the paper well documented and is the documentation provided alongside the assets?
    \item[] Answer: \answerYes{}.
    \item[] Justification: The paper introduces PanoV2V-15K, a new paired clean--degraded panoramic video dataset. We document its construction protocol, scene composition, trajectory generation, clean/degraded pair generation, evaluation splits, intended research use, and license terms. Additional dataset statistics and examples are provided in the appendix. Dataset documentation and access instructions are provided alongside the released assets.
    \item[] Guidelines:
    \begin{itemize}
        \item The answer \answerNA{} means that the paper does not release new assets.
        \item Researchers should communicate the details of the dataset\slash code\slash model as part of their submissions via structured templates.
        \item The paper should discuss whether and how consent was obtained from people whose asset is used.
        \item At submission time, remember to anonymize your assets if applicable.
    \end{itemize}

\item {\bf Crowdsourcing and research with human subjects}
    \item[] Question: For crowdsourcing experiments and research with human subjects, does the paper include the full text of instructions given to participants and screenshots, if applicable, as well as details about compensation?
    \item[] Answer: \answerNA{}.
    \item[] Justification: The paper does not involve crowdsourcing or human-subject research. The dataset is constructed from synthetic indoor scene assets, and any captions or annotations are automatically generated rather than collected from human participants.
    \item[] Guidelines:
    \begin{itemize}
        \item The answer \answerNA{} means that the paper does not involve crowdsourcing nor research with human subjects.
        \item Including this information in the supplemental material is fine, but if the main contribution involves human subjects, then as much detail as possible should be included in the main paper.
        \item According to the NeurIPS Code of Ethics, workers involved in data collection or curation should be paid at least the minimum wage in the country of the data collector.
    \end{itemize}

\item {\bf Institutional review board (IRB) approvals or equivalent for research with human subjects}
    \item[] Question: Does the paper describe potential risks incurred by study participants, whether such risks were disclosed to the subjects, and whether Institutional Review Board approvals or equivalent review were obtained?
    \item[] Answer: \answerNA{}.
    \item[] Justification: The paper does not involve human-subject research, crowdsourcing, or collection of personal data, so IRB approval or equivalent review is not applicable.
    \item[] Guidelines:
    \begin{itemize}
        \item The answer \answerNA{} means that the paper does not involve crowdsourcing nor research with human subjects.
        \item Depending on the country in which research is conducted, IRB approval or equivalent may be required for any human subjects research.
        \item For initial submissions, do not include any information that would break anonymity.
    \end{itemize}

\item {\bf Declaration of LLM usage}
    \item[] Question: Does the paper describe the usage of LLMs if it is an important, original, or non-standard component of the core methods in this research?
    \item[] Answer: \answerNA{}.
    \item[] Justification: LLMs are not used as an important, original, or non-standard component of the core Rein3D method. The panoramic video restoration model and 3DGS refinement pipeline do not depend on LLM-generated outputs. If automatically generated scene captions are included as optional dataset metadata, they are not used for training the panoramic restoration model or for the main 3D reconstruction pipeline.
    \item[] Guidelines:
    \begin{itemize}
        \item The answer \answerNA{} means that the core method development in this research does not involve LLMs as any important, original, or non-standard components.
        \item Please refer to the NeurIPS LLM policy for what should or should not be described.
    \end{itemize}

\end{enumerate}
\end{document}